\documentclass[letterpaper, 10 pt, conference]{ieeeconf}  
\IEEEoverridecommandlockouts
\usepackage{cite}
\usepackage{amsmath,amssymb,amsfonts}
\usepackage{algorithmic}
\usepackage{graphicx}
\usepackage{textcomp}
\usepackage{color}
\usepackage[table]{xcolor}
\usepackage{colortbl}
\usepackage[colorlinks,linkcolor=blue,anchorcolor=blue,citecolor=blue]{hyperref}

\def\BibTeX{{\rm B\kern-.05em{\sc i\kern-.025em b}\kern-.08em
    T\kern-.1667em\lower.7ex\hbox{E}\kern-.125emX}} 

\title{\LARGE \bf UAV-DETR: Efficient End-to-End Object Detection for Unmanned Aerial Vehicle Imagery}

\author{Huaxiang Zhang, Hao Zhang, Kai Liu, Zhongxue Gan*, and Guo-Niu Zhu*,~\IEEEmembership{Member,~IEEE}
\thanks{*This work is supported by Shanghai Municipal Science and Technology Major Project under Grant 2021SHZDZX0103 and Key Project of Comprehensive Prosperity Plan of Fudan University under Grant XM06231744. (Corresponding author: Zhongxue Gan and Guo-Niu Zhu)}
\thanks{All authors are with the Academy for Engineering and Technology, Fudan University, Shanghai 200433, China (e-mail: guoniu\_zhu@fudan.edu.cn).}
}

\begin{document}
\maketitle
\thispagestyle{empty}
\pagestyle{empty}

\begin{abstract}
Unmanned aerial vehicle object detection (UAV-OD) has been widely used in various scenarios. However, most existing UAV-OD algorithms rely on manually designed components, which require extensive tuning. End-to-end models that do not depend on such manually designed components are mainly designed for natural images, which are less effective for UAV imagery. To address such challenges, this paper proposes an efficient detection transformer (DETR) framework tailored for UAV imagery, i.e., UAV-DETR. The framework includes a multi-scale feature fusion with frequency enhancement module, which captures both spatial and frequency information at different scales. In addition, a frequency-focused downsampling module is presented to retain critical spatial details during downsampling. A semantic alignment and calibration module is developed to align and fuse features from different fusion paths. Experimental results demonstrate the effectiveness and generalization of our approach across various UAV imagery datasets. On the VisDrone dataset, our method improves AP by 3.1\% and \(\text{AP}_{50}\) by 4.2\% over the baseline. Similar enhancements are observed on the UAVVaste dataset. The project page: \url{https://github.com/ValiantDiligent/UAV-DETR}
\end{abstract}

\section{Introduction}
Camera-equipped unmanned aerial vehicles (UAVs) have been widely applied in various fields \cite{10611308}. As one of the core technologies in these applications, UAV object detection (UAV-OD) has attracted considerable attention \cite{hicyolo}. Popular UAV-OD algorithms often rely on manually designed components, such as non-maximum suppression (NMS) and anchor boxes generated based on human expertise \cite{TONG2020103910}. These components require extensive tuning for different tasks, which are complex and inefficient in practical applications. In contrast, end-to-end models are free from these issues. Therefore, it would be a good choice to develop end-to-end models for UAV-OD.
\begin{figure}[ht]
\centering
\includegraphics[width=8.5cm]{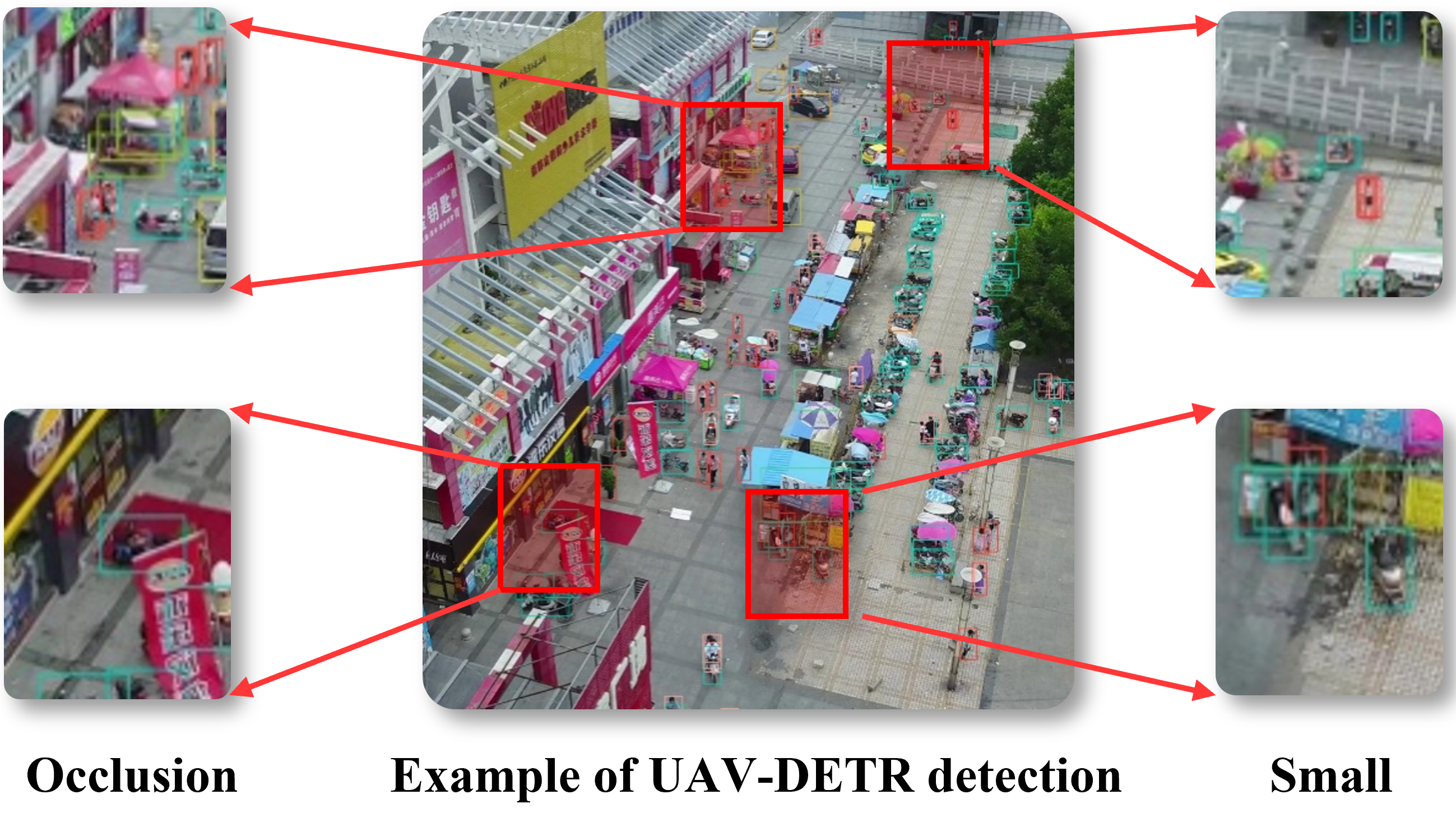}
\caption{Challenges in UAV-OD}
\label{figure_1}
\end{figure}

As a popular end-to-end model, the Detection Transformer (DETR) \cite{detr} utilizes the transformer architecture to create an end-to-end detector. Recent studies have improved the small object detection capabilities of DETR models, but their high computational cost and poor real-time performance make them unsuitable for real-time scenarios \cite{zhudeformable}. To tackle such issues, Zhao et al. \cite{Zhao_2024_CVPR} introduced a real-time detection transformer (RT-DETR), which surpassed the popular you only look once (YOLO) framework in both accuracy and speed. However, existing DETRs are predominantly designed for natural images, which poses challenges when applied to UAV image analysis.

As shown in Fig. \ref{figure_1}, the object features in UAV vision are more complex than those in normal vision. Aerial imagery suffers from challenges such as small object sizes and occlusion. Therefore, detecting objects in UAV images typically benefits from detailed feature extraction \cite{hicyolo}. In cases where local features may not provide sufficient information, incorporating the relationship between the object and its surrounding environment would be a viable option to enhance the detection accuracy \cite{TONG2020103910, Li_2023_ICCV}.

To handle the challenge of object detection in aerial imagery, this paper proposes an efficient detection transformer framework for UAV imagery, namely UAV-DETR. We enhance our model by leveraging both spatial and frequency domain information across multiple scales to retain high-frequency components. We present a frequency-focused downsampling strategy to preserve critical spatial details during downsampling. Finally, we enhance the semantic representation capability of the model by aligning features from different feature fusion paths.

Our main contributions are summarized as follows.
\begin{enumerate}
\item We propose UAV-DETR, an efficient end-to-end detector transformer for UAV imagery. The framework achieves superior accuracy and real-time performance. We put forward three models of different sizes to meet various performance requirements.
\item We present a multi-scale feature fusion with frequency enhancement module to enhance the detection of small and occluded objects.
\item We develop a frequency-focused downsampling module that retains dual-domain information.
\item We propose a semantic alignment and calibration module to align features that from different feature fusion paths to boost detection performance.
\end{enumerate}

\section{Related Work}
\subsection{Object Detection in UAV Imagery}
Object detection in UAV imagery presents unique challenges, particularly in detecting small objects and managing occlusions. Moreover, UAV-OD often needs to be deployed on hardware platforms, which requires to balance between real-time performance and computational complexity \cite{TONG2020103910}. Some studies \cite{Yang_2022_CVPR, yang2019clustered} presented a coarse-to-fine processing pipeline for UAV-OD. Although these two-stage methods achieve high accuracy, they introduce significant computational overhead \cite{feng2023lightweight}, which makes them unsuitable for resource-limited environments. To address this issue, another strategy focused on developing optimized single-stage models to balance between detection accuracy and efficiency. Meanwhile, lots of works \cite{hicyolo, MITTAL2020104046} suggested to capture more features relevant for detecting small objects, with most of them concentrating on utilizing higher-resolution feature maps. In addition, some methods \cite{shen2023multiple, liu2021survey} leveraged contextual information to enhance small object detection. 

Among these studies, most UAV-OD methods are devoted to lightweighting models or optimizing the processing pipeline for practical use. There is limited research on post-processing techniques. Additionally, these methods mainly extract detailed features and contextual information in the spatial domain, while the frequency domain is underutilized.

\subsection{Real-Time End-to-End Object Detection}
Many single-stage UAV-OD models are based on the YOLO series models due to their balance of performance and real-time capability \cite{MITTAL2020104046, liu2021survey}. However, these detectors typically require NMS for post-processing, which not only slows down inference but also introduces hyperparameters that can lead to instability in speed and accuracy. 

By contrast, RT-DETR \cite{Zhao_2024_CVPR} is the first real-time and end-to-end object detector, which eliminates the influence of NMS. It surpasses even the most powerful YOLO models in both speed and accuracy through attention-based intra-scale feature interaction, CNN-based cross-scale feature fusion, and uncertainty-minimal query selection.

RT-DETR leverages global attention mechanisms to capture long-range dependencies and contextual information, which make it more flexible and effective. Its end-to-end design strategy enables it more suitable for deployment on UAV platforms compared to the YOLO series models. 

\subsection{Feature Fusion}
Feature fusion techniques aim to combine multi-scale feature maps to improve object detection. However, semantic gaps between features at different levels pose challenges, especially for detecting small and densely distributed objects\cite{deng2020feature}. An intuitive approach is to sum or concatenate feature maps from different layers, though it often leads to spatial feature misalignment. For example, Li et al. \cite{CSFCN} presented pooling-based and sampling-based attention mechanisms to investigate these issues.

While these methods focus on spatial feature fusion, the frequency-domain information is not taken into consideration. Although some works\cite{cui2024omni},\cite{FTMF} explored frequency-domain fusion, they fall short in effective multi-scale fusion across both spatial and frequency domains.

In contrast, our UAV-DETR, a single-stage model with a DETR-like architecture, performs multi-scale feature fusion in both spatial and frequency domains. By using learned offsets to align features across different fusion paths, our approach addresses the misalignment issue and enhances detection performance.

\section{Methodology}
\begin{figure*}[ht]
\centering
\includegraphics[width=17.5cm]{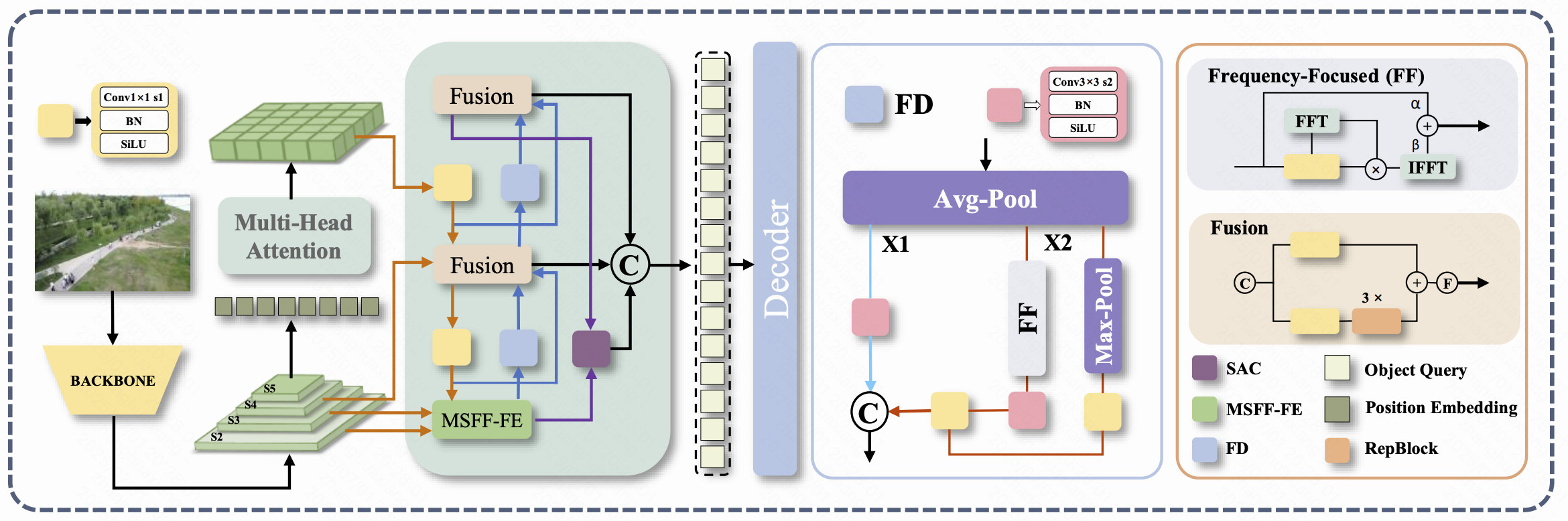}
\caption{Overview of the UAV-DETR. MSFF-FE represents the Multi-Scale Feature Fusion with Frequency Enhancement module; FD denotes Frequency-focused Downsampling; SAC is Semantic Alignment and Calibration. FFT and IFFT denote fast Fourier transform and its inverse operation, respectively.} 
\label{figure_overview}
\end{figure*}

As shown in Fig. \ref{figure_overview}, this study proposes a UAV-DETR, which is built upon the architecture of RT-DETR \cite{Zhao_2024_CVPR}. We enhance the model with three components, i.e., multi-scale feature fusion with frequency enhancement, frequency-focused downsampling, and semantic alignment and calibration. Additionally, we introduce inner Scylla intersection over union (Inner-SIoU) \cite{gevorgyan2022siou, zhang2023inner} to replace the generalized intersection over union (GIoU).

\subsection{Multi-Scale Feature Fusion with Frequency Enhancement}
In traditional feature fusion, high-frequency components are often lost. In this study, we present a multi-scale feature fusion with frequency enhancement module (MSFF-FE) to preserve small object details by combining spatial and frequency domain information across multiple scales.
\begin{figure}[ht]
\centering
\includegraphics [width=8.5cm]{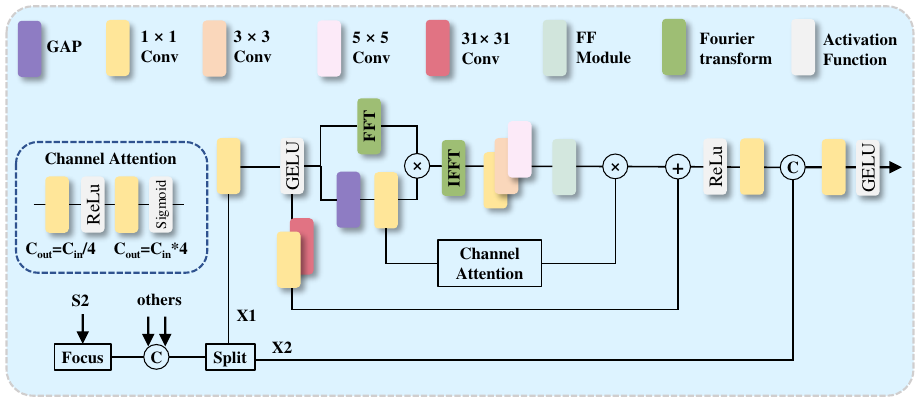}
\caption{Structure of multi-scale feature fusion with frequency enhancement. By default, the number of output channels in the convolutional layers equals the number of input channels. The structural diagram of the Frequency-Focused module is included in Fig. \ref{figure_overview}. }
\label{figure_msff}
\end{figure}

In the proposed MSFF-FE architecture, the Focus module \cite{yolov8_ultralytics} is employed to compress spatial information from satellite imagery into channel dimensions through stride-based feature slicing and a convolution layer. As depicted in Fig. \ref{figure_msff}, we process S2 through Focus modules and concatenate the resulting feature map with other input feature maps along the channel dimension as input. The MSFF-FE adopts a cross-stage partial strategy \cite{Wang_2020_CVPR_Workshops}. It partitions the input feature map \(x \in \mathbb{R}^{C \times H \times W}\) into two parts: \(x_1 \in \mathbb{R}^{C_1 \times H \times W}\) and \(x_2 \in \mathbb{R}^{C_2 \times H \times W}\). To avoid excessive computational cost, we set $C_1 = \frac{C}{4}$ and $C_2 = \frac{3C}{4}$. The first part \(x_1\) undergoes multi-scale and frequency enhancement, while the second part \(x_2\) is concatenated with the processed features from \(x_1\), followed by a $1 \times 1$ convolution to fuse the two branches. Firstly, the feature map \(x_1\) passes through a $1 \times 1$ convolution to adjust its channel dimensions, followed by the GELU activation function\cite{hendrycks2016gaussian} for non-linearity, resulting in \(x_{\text{conv}}\). As shown in Eq. \eqref{eq1}, we transform \(x_{\text{conv}}\) into the frequency domain by Fourier transform (F). To refine the frequency-domain features, we design a Global Average Pooling (GAP) layer, followed by a convolutional layer. After refinement, the features are transformed back to the spatial domain via inverse Fourier transform (IF).
\begin{equation}
x_{\text{sp}} = \left| IF\left( \text{Conv}_{1 \times 1}(\text{GAP}(x_{\text{conv}})) \cdot F(x_{\text{conv}}) \right) \right|
\label{eq1}
\end{equation}

To capture multi-scale information, three convolutions with different kernel sizes are applied to \(x_{\text{sp}}\).
\begin{equation}
x_{\text{sc}} = \text{Conv}_{1 \times 1}(x_{\text{sp}}) + \text{Conv}_{3 \times 3}(x_{\text{sp}}) + \text{Conv}_{5 \times 5}(x_{\text{sp}})
\end{equation}

In our design, we also leverage a channel attention mechanism on \( \text{GAP}(x_{\text{conv}}) \) to further refine the \(x_{\text{sc}}\). Then, We leverage the gating mechanism to modulate and refine the multi-scale features.
\begin{equation}
x_{\text{F}} = \alpha \cdot IF(F(\text{Conv}_{1 \times 1}(x_{\text{sc}})) \cdot \text{Conv}_{1 \times 1}(x_{\text{sc}})) + \beta \cdot x_{\text{sc}}
\end{equation}
where \(\alpha\) and \(\beta\) are learned parameters balancing the spatial and frequency components. We refer to this equation as the Frequency-Focused (FF) module, which will be used in later stages of the network. The FF module dynamically learns and adjusts filtering coefficients in the frequency domain through $1 \times 1$ convolution layers. Additionally, we incorporate residual connections. This design preserves the original spatial features. At the same time, it allows for controllable fusion of frequency-enhanced components. Before combining with the unprocessed \(x_2\), the enhanced features undergo final fusion.
\begin{equation}
x_{\text{final}} = x_1 + \text{Conv}_{31 \times 31}(x_{\text{conv}}) + \text{Conv}_{1 \times 1}(x_{\text{conv}}) + x_{\text{F}}
\end{equation}

We employ large convolution kernels to capture long-range dependencies, while small convolution kernels are utilized to optimize channel-wise information. Additionally, residual connections are incorporated to accelerate model training. The final output is obtained by concatenating \(x_{\text{final}}\) with \(x_2\), followed by a $1 \times 1$ convolution and GELU activation function.

\subsection{Frequency-Focused Downsampling }
As illustrated in Fig. \ref{figure_overview}, in the frequency-focused downsampling (FD) module, the input feature map \(x \in \mathbb{R}^{C \times H \times W}\) is first downsampled using average pooling with a kernel size of 2 and a stride of 1. As a result, the pooled feature map \(x_p\) is obtained and then it is divided into two parts along the channel dimension: \(x_1\) and \(x_2\). Each of them are processed in parallel.

\(x_1\) is processed using a $3 \times 3$ convolution with stride 2 and padding 1, which reduces its spatial dimensions while preserving key features. Then, \(x_1'\) is obtained. \(x_2\) undergoes parallel processing, where one path applies FF module to enhance important feature components. Accordingly, \(x_f\) is obtained. The other path applies max pooling with a $3 \times 3$ kernel and stride 2, followed by a $1 \times 1$ convolution to reduce the number of channels. Correspondingly, we get \(x_p'\). The two outputs, \(x_f\) and \(x_p'\), are concatenated along the channel dimension and passed through a $1 \times 1$ convolution to reduce the number of channels to the desired size, which results in \(x_2'\). Finally, \(x_1'\) and \(x_2'\) are concatenated to form the final output of the module.
\subsection{Semantic Alignment and Calibration}
\begin{figure}[ht]
\centering
\includegraphics [width=8.5cm]{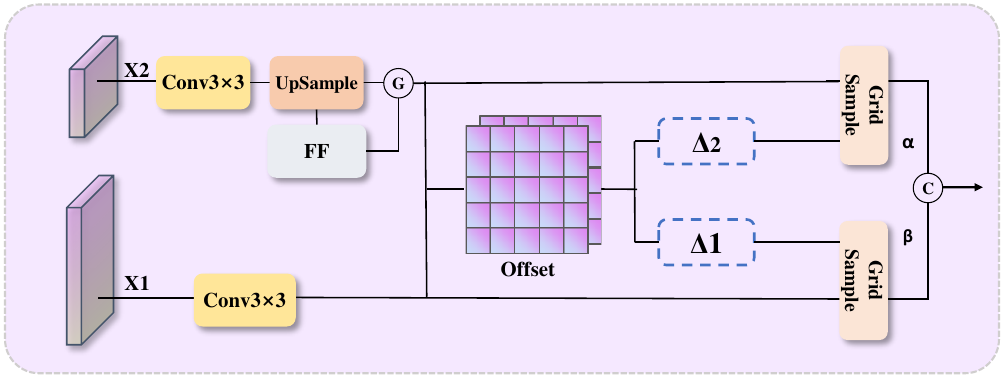}
\caption{Structure of semantic alignment and calibration.}
\label{figure_SAC}
\end{figure}

As shown in Fig. \ref{figure_SAC}, the semantic alignment and calibration (SAC) module is designed to fuse and align features obtained from different fusion processes.

Given two input features \(x_1 \in \mathbb{R}^{C_1 \times H_1 \times W_1}\) and \(x_2 \in \mathbb{R}^{C_2 \times H_2 \times W_2}\), the SAC module first unifies the number of channels to a common dimension \(C\) through separate convolution layers. Then, the feature \(x_2\) is upsampled using bilinear interpolation to match the spatial dimensions of \(x_1\). To enhance feature \(x_2\), we apply FF module that generates a frequency-enhanced feature \(x_{\text{freq}}\). Afterward, we fuse the frequency-enhanced feature with the original upsampled feature \(x_2\). A gating mechanism is employed to balance the contributions from both the spatial and frequency domains.
\begin{equation}
x_{\text{fused}} = G(x_2) \cdot x_{\text{freq}} + (1 - G(x_2)) \cdot x_2
\end{equation}
where \(G\) is a learned gating function, which is used to ensure an adaptive fusion of spatial and frequency-domain information.

To address misalignment between \(x_1\) and \(x_{\text{fused}}\), the SAC module learns 2D offsets \(\Delta_1\) and \(\Delta_2\), which adjust the sampling grid for each feature map. These offsets are generated through a convolution layer. Using the learned offsets, we adjust the spatial coordinates of the features through a grid-based sampling operation \cite{jaderberg2015spatial} to ensure the alignment of both features.
\begin{equation}
x_1^{\text{aligned}} = \text{GridSample}(x_1, \Delta_1), 
\end{equation}
\begin{equation}
x_{\text{fused}}^{\text{aligned}} = \text{GridSample}(x_{\text{fused}}, \Delta_2)
\end{equation}

GridSample employs learnable offsets $\Delta$ to dynamically warp feature coordinates via bilinear interpolation to enable geometric alignment of multi-scale features. This differentiable formulation preserves gradient flow while adapting to viewpoint variations and scale discrepancies in UAV imagery.

Then, an element-wise weighted summation is used to fuse the aligned features, where \(\alpha\) and \(\beta\) are learned attention weights that balance the contributions from each aligned feature.
\begin{equation}
x_{\text{output}} = \alpha \cdot x_1^{\text{aligned}} + \beta \cdot x_{\text{fused}}^{\text{aligned}}
\end{equation}

\subsection{LOSS Function }
RT-DETR uses GIoU loss for bounding box regression, which is less effective for small object detection, especially when the Intersection over Union (IoU) values are low. To address this issue, we adopt Inner-SIoU, which enhances both small object detection and geometric alignment. Inner-SIoU combines Inner-IoU\cite{zhang2023inner} and SCYLLA-IoU (SIoU)\cite{gevorgyan2022siou}. It scales the auxiliary bounding box by 1.25 to improve sensitivity and speed up convergence. The InnerIoU is defined using expanded auxiliary bounding boxes. The Inner-IoU is calculated as:
\begin{equation}
\text{Inner-IoU} = \frac{|B^{\text{inner}} \cap B^{\text{inner}}_{\text{gt}}|}{|B^{\text{inner}} \cup B^{\text{inner}}_{\text{gt}}|}
\end{equation}
where \(B^{\text{inner}}\) and \(B^{\text{inner}}_{\text{gt}}\) represent the expanded predicted and ground truth boxes, respectively. The width and height of both boxes are scaled by a factor of \(1.25\). The terms \(|B^{\text{inner}} \cap B^{\text{inner}}_{\text{gt}}|\) and \(|B^{\text{inner}} \cup B^{\text{inner}}_{\text{gt}}|\) denote the area of overlap and the union area between the expanded boxes, respectively.

The Inner-SIoU loss is defined as:
\begin{equation}
L_{\text{Inner-SIoU}} = L_{\text{SIoU}} + \text{IoU} - \text{Inner-SIoU}
\end{equation}
where \( \text{IoU} \) is the standard IoU loss, and \( L_{\text{SIoU}} \) includes angle, distance, and shape penalties.

\section{Experiments}
\subsection{Experimental Setup}
\textbf{Datasets:}
We conduct quantitative experiments on two object detection datasets: VisDrone\cite{zhu2021detection} and UAVVaste\cite{rs13050965}. The VisDrone-2019-DET dataset comprises 6,471 training images, 548 validation images, and 3,190 test images, all captured from drones at varying altitudes across different locations. Each image is annotated with bounding boxes for ten predefined object categories: pedestrian, person, car, van, bus, truck, motorbike, bicycle, awning-tricycle, and tricycle. We use the VisDrone-2019-DET training set and validation set for training and testing, respectively.

In addition, we further train the UAV-DETR network with UAVVaste dataset to validate the capacity to generalize across datasets. UAVVaste is a dataset designed specifically for aerial rubbish detection. It consists of 772 images and 3716 hand-labeled annotations of waste in urban and natural environments such as streets, parks, and lawns. We choose the training set for training and the test set for testing. 

\textbf{Implementation Details:}
All models are trained on NVIDIA GeForce RTX 3090. The UAV-DETR model is based on RT-DETR\cite{Zhao_2024_CVPR}, with three variations: one using ResNet18, another with ResNet50, and a third incorporating EfficientFormerV2\cite{li2022rethinking} as the backbone. We implement our approach and train the framework for 400 epochs with a batch size of 4. We use an early stopping mechanism with a patience setting of 20. The UAV-DETR network is optimized using AdamW\cite{loshchilov2017decoupled} with a learning rate of 0.0001, a momentum of 0.9. We scale our input images to 640$\times$640 pixels and use the data augmentation methods from the RT-DETR model to ensure consistency across experiments. Additionally, we apply mixup \cite{zhang2018mixup} and Mosaic \cite{bochkovskiy2020yolov4} techniques, with Mosaic set to a probability of 1 and mixup set to a probability of 0.2. We report the standard COCO metrics, including AP (averaged over uniformly sampled IoU thresholds ranging from 0.50-0.95 with a step size of 0.05), AP$_{50}$, as well as AP at different scales: AP$_{S}$, AP$_{M}$.
\begin{table*}[htbp]
\caption{Experimental Results on the VisDrone-2019-DET Dataset}
\begin{center}
\footnotesize
\begin{tabular}{l c c c c c c} 
\hline
\textbf{Model} & \textbf{Publication} & \textbf{InputSize} & \textbf{Params(M)} & \textbf{GFLOPs} & \textbf{AP} & \textbf{AP$_{50}$} \\
\hline

\multicolumn{7}{l}{\cellcolor{gray!20}\textit{Real-time Object Detectors}} \\ 
YOLOv8-M\cite{yolov8_ultralytics} & - & 640×640& 25.9 & 78.9 & 24.6 & 40.7 \\
YOLOv8-L\cite{yolov8_ultralytics} & - & 640×640& 43.7 & 165.2 & 26.1 & 42.7 \\
YOLOv9-S\cite{wang2024yolov9} & ECCV2025 & 640×640& 7.2 & 26.7 & 22.7 & 38.3 \\
YOLOv9-M\cite{wang2024yolov9} & ECCV2025 & 640×640& 20.1 & 76.8 & 25.2 & 42.0 \\
YOLOv10-M\cite{wang2024yolov10} & arXiv2024& 640×640& 15.4 & 59.1 & 24.5 & 40.5 \\
YOLOv10-L\cite{wang2024yolov10} & arXiv2024& 640×640& 24.4 & 120.3 & 26.3 & 43.1 \\
YOLOv11-S\cite{khanam2024yolov11} & arXiv2024 & 640×640 & 9.4 & 21.3 & 23.0 & 38.7 \\
YOLOv11-M\cite{khanam2024yolov11} & arXiv2024 & 640×640 & 20.0 & 67.7 & 25.9 & 43.1 \\
\hline

\multicolumn{7}{l}{\cellcolor{gray!20}\textit{Object Detectors for UAV Imagery}} \\ 
PP-YOLOE-P2-Alpha-l\cite{ppdet2019} & - & 640×640& 54.1& 111.4& 30.1 & 48.9 \\
QueryDet\cite{Yang_2022_CVPR}& CVPR2022 & 2400×2400 & 33.9& 212& 28.3& 48.1 \\
ClusDet\cite{yang2019clustered}& ICCV2019 & 1000×600 & 30.2& 207& 26.7& 50.6 \\
DCFL\cite{Xu_2023_CVPR}& CVPR2023 & 1024×1024& 36.1& 157.8& -& 32.1 \\
HIC-YOLOv5\cite{hicyolo} & ICRA2024 & 640×640& 9.4& 31.2& 26.0& 44.3\\
\hline
\multicolumn{7}{l}{\cellcolor{gray!20}\textit{End-to-end Object Detectors}} \\ 
DETR\cite{detr} & ECCV2020 & 1333×750& 60& 187& 24.1& 40.1 \\
Deformable DETR\cite{zhudeformable} & ICLR2020 & 1333×800& 40& 173& 27.1& 42.2 \\
Sparse DETR\cite{roh2022sparse}  & ICLR2022 & 1333×800& 40.9& 121& 27.3& 42.5 \\
RT-DETR-R18\cite{Zhao_2024_CVPR} & CVPR2024& 640×640& 20& 60.0& 26.7& 44.6 \\
RT-DETR-R50\cite{Zhao_2024_CVPR} & CVPR2024& 640×640& 42& 136& 28.4& 47.0 \\
\hline

\multicolumn{7}{l}{\cellcolor{gray!20}\textit{Real-time End-to-end Object Detectors for UAV Imagery}} \\ 
UAV-DETR-EV2(Ours) & -& 640×640 & 13 & 43 & 28.7& 47.5 \\
UAV-DETR-R18(Ours) & -& 640×640 & 20 & 77 & 29.8& 48.8 \\
UAV-DETR-R50(Ours) & - & 640×640 & 42 & 170 & \textbf{31.5} & \textbf{51.1} \\
\hline
\end{tabular}%
\end{center}
\label{table:detectors}
\end{table*}
\begin{table*}[htbp]
\caption{Experimental Results on UAVASTE Dataset}
\centering
\begin{tabular}{lcccccc}
\hline
\textbf{Model} & \textbf{Params (M)} & \textbf{GFLOPs} & \textbf{AP$_{S}$} & \textbf{AP$_{M}$} & \textbf{AP} & \textbf{AP$_{50}$}\\
\hline
YOLOv11-S\cite{khanam2024yolov11} & 9.4 & 21.3 & 27.3 & 48.7 & 27.8 & 63.0\\
HIC-YOLOv5\cite{hicyolo} & 9.4 & 31.2 & 30.8 & 20.9 & 30.5 & 65.1\\
RT-DETR-R18\cite{Zhao_2024_CVPR} & 20.0 & 57.3 & 35.8 & 64.8 & 36.3 & 72.6\\
RT-DETR-R50\cite{Zhao_2024_CVPR} & 42.0 & 129.9 & 37.0 & 62.3 & 37.4 & 73.5\\
\hline
UAV-DETR-EV2 (Ours)& 13 & 43 & 36.7 & 63.3 & 37.1 & 70.6\\
UAV-DETR-R18 (Ours)& 20 & 77 & 36.6 & 64.6 & 37.0 & 74.0\\
UAV-DETR-R50 (Ours)& 42 & 170 & 37.1 & 61.3 & 37.5 & 75.9\\
\hline
\end{tabular}
\label{table_vaste}
\end{table*}
\subsection{Comparative Experiments} 
As listed in Table \ref{table:detectors}, on the VisDrone dataset, our UAV-DETR-R18  achieves a 3.1\% improvement in \(\text{AP}\), a 4.2\% increase in \(\text{AP}_{50}\) compared to the baseline RT-DETR-R18. Similarly, the UAV-DETR-R50 sees a 3.1\% increase in \(\text{AP}\), a 4.1\% rise in \(\text{AP}_{50}\) compared to the baseline. UAV-DETR-EV2 also demonstrates an accuracy advantage over HIC-YOLOv5, which has a similar computational cost. UAV-DETR-R18 outperforms all methods with a computational cost below 100 GFLOPs, which achieves the best accuracy in its class. Furthermore, we compared our method with other object detectors that have computational costs similar to UAV-DETR-R50, and the results show that our approach also outperforms the others in terms of accuracy. Remarkably, our method still exhibits outstanding performance, even compared to approaches like PP-YOLOE-P2-Alpha-l \cite{ppdet2019}, which typically benefit from extensive pre-training. Such comparisons are often considered unfair due to the significant advantages conferred by extensive pre-training. To further demonstrate the generalizability of UAV-DETR, we also evaluated the method on the UAVVaste dataset. The results are given in Table \ref{table_vaste}. Notably, UAV-DETR still maintains a competitive advantage compared to other models. Compared to VisDrone, the UAVVaste dataset has a smaller amount of data. The strong performance of our model demonstrates that it does not rely on large amounts of annotated data.

 \begin{table}[htbp]
\caption{Results of the Ablation Study.}
\centering
\begin{tabular}{cccccccc}
\hline
\textbf{ Baseline} & \textbf{ IS} & \textbf{MSFF-FE} & \textbf{ FD} & \textbf{ SAC} & \textbf{ AP} & \textbf{ AP$_{50}$} \\
\hline
\checkmark &      &      &      &      & 26.7 & 44.6 \\
\checkmark & \checkmark &      &      &      & 27.1 & 45.3 \\
\checkmark & \checkmark & \checkmark &      &      & 28.4 & 46.9\\
\checkmark & \checkmark & \checkmark & \checkmark &      & 28.4 & 47.1 \\
\checkmark & \checkmark & \checkmark & \checkmark & \checkmark & 29.8 & 48.8 \\
\hline
\end{tabular}
\label{tab:performance_comparison}
\end{table}
\begin{table}[htbp]
\caption{Comparison of IoU Metrics}
\centering
\begin{tabular}{lcc}
\hline
\textbf{IoU} & \textbf{AP} & \textbf{AP$_{50}$} \\
\hline
GIoU & 29.0& 48.4\\
Inner-SIoU (ratio=1.20) & 29.5  & 48.7  \\
Inner-SIoU (ratio=1.25) & 29.8   & 48.8  \\
Inner-SIoU (ratio=1.30) & 29.3  & 48.6 \\
\hline
\end{tabular}
\label{table_iou}
\end{table}
\begin{table}[htbp]
\caption{Comparison of Model Performance Metrics}
\centering
\begin{tabular}{lccc}
\hline
\textbf{Model} & \textbf{Params(M)}& \textbf{GFLOPs}& \textbf{FPS} \\
\hline
RT-DETR-R50\cite{Zhao_2024_CVPR}& 42& 130  & 89  \\
RT-DETR-R18\cite{Zhao_2024_CVPR}& 20& 60 & 183  \\
UAV-DETR-R50& 42  & 170  & 65   \\
UAV-DETR-R18& 20  & 77  & 124  \\
UAV-DETR-EV2& 13  & 43  & 116  \\
\hline
\end{tabular}
\label{table_FPS}
\end{table}
\subsection{Ablation Studies}
We conducted ablation experiments on the VisDrone dataset using UAV-DETR-R18 to analyze the impact of each component on detection accuracy. Table \ref{tab:performance_comparison} presents the performance comparison with different configurations, where IS stands for Inner-SIoU, MSFF-FE represents Multi-Scale Feature Fusion with Frequency Enhancement module, FD stands for Frequency-Focused Downsampling module, and SAC refers to Semantic Alignment and Calibration module.

The baseline RT-DETR-R18 achieves an \text{AP} of 26.7 and \text{AP$_{50}$} of 44.6. After incorporating Inner-SIoU, \text{AP} increased to 27.1, which shows that improving the loss function positively impacts performance. Adding the MSFF-FE module further improves the \text{AP} to 28.4, which demonstrates the benefits of incorporating multi-scale feature fusion and frequency enhancement. The addition of FD improves AP$_{50}$ to 47.1. When all components are combined, UAV-DETR-R18 achieves the highest performance with an \text{AP} of 29.8 and an \text{AP$_{50}$} of 48.8, which showcases the cumulative impact of each module on detection accuracy. As given in Table \ref{table_iou}, our experiments demonstrate that setting the ratio of Inner-SIOU to 1.25 is an appropriate choice. Lastly, we calculated the Frames Per Second (FPS) for both RT-DETR models and UAV-DETR models using the PyTorch implementation in 32-bit Floating Point Precision, as shown in Table \ref{table_FPS}. The result demonstrates that UAV-DETR largely maintains the real-time performance of RT-DETR.
\begin{figure}[ht]
\centering
\includegraphics[width=8.5cm]{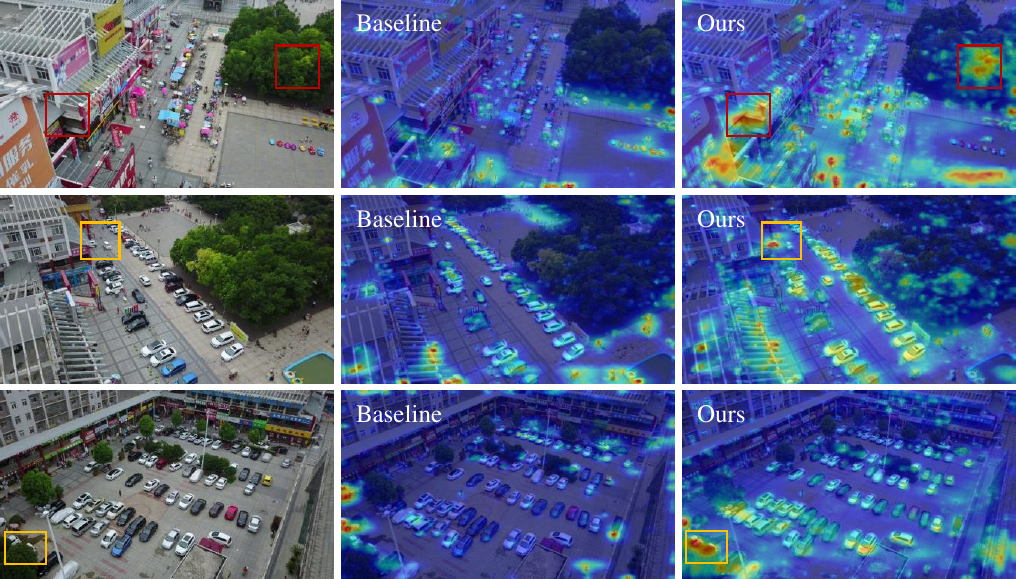}
\caption{The heatmap of RT-DETR-R18 and UAV-DETR-R18. The brighter areas in the heatmap indicate stronger attention by the model. Our model shows greater focus on small objects and their surrounding environment. The yellow boxes highlight areas where our model performs better in detecting occluded objects. The red boxes indicate regions where the model misfocuses on noise.} 
\label{figure_heatmap}
\end{figure}

\subsection{Visualization}
In Fig. \ref{figure_heatmap}, we present heatmaps for small objects in the VisDrone dataset, focusing on backpropagation through the predictions of the bounding box. Compared to the baseline model, UAV-DETR demonstrates a notably improved ability to localize small objects. In the heatmaps of our model, small objects exhibit higher heat values, which indicates that the model can more effectively capture the features of these small objects. Additionally, it can be observed that UAV-DETR pays more attention to the surrounding information of small objects, which demonstrates the model's ability to better utilize contextual information during detection. As a result, as shown in the yellow box in Fig. \ref{figure_heatmap}, UAV-DETR also performs well in localizing occluded objects. Therefore, UAV-DETR does not severely degrade the baseline's real-time performance. After deployment on different hardware platforms, FPS can obtain significant improvements\cite{performance}.

\subsection{Discussion}
Compared to other UAV-OD models, UAV-DETR has two key differences. First, it eliminates the need for NMS and anchor setting, which greatly reduces the complexity of model deployment. Second, UAV-DETR leverages dual-domain information during feature fusion. As shown in Table \ref{table:detectors}, our model achieves higher accuracy compared to other detectors with similar computational costs. 

There are several reasons that contribute to this improved performance. Firstly, our model retains more high-frequency features, which are crucial for detecting small objects. In traditional feature fusion and downsampling processes, high-frequency features are often lost, which makes it harder to recover edge and texture details. These details are particularly important for UAV-OD. To address this, we introduced the MSFF-FE and FD modules, which enable the model to combine spatial and frequency domain information during feature fusion and downsampling to eusure that important high-frequency components are preserved. 

Secondly, our model makes better use of contextual information. When small objects are difficult to detect based on fine details, their surrounding semantic context becomes crucial. The operations in the frequency domain help the model capture global patterns, which improves the detection accuracy. However, these operations can sometimes lead to misalignment between the semantic and spatial information of different feature maps. To handle this problem, we designed the SAC module, which aligns features from different fusion paths and enhances overall detection performance. The ablation study in Table \ref{tab:performance_comparison} demonstrates the effectiveness of these modules.

Our findings suggest that utilizing frequency domain information can enhance the performance of UAV-OD. We hope this approach will provide insights into how frequency information can be better used in UAV-OD tasks. 
However, as shown in the red box in Fig. \ref{figure_heatmap}, the model occasionally focuses on irrelevant regions, which presents a challenge that we aim to address in future work.  

\section{Conclusion}
This paper proposes UAV-DETR, a real-time end-to-end object detector specifically designed for UAV imagery. By developing the MSFF-FE module, FD module, and SAC module, UAV-DETR helps alleviate the difficulties of detecting small and occluded objects in aerial images. Each of our modules plays a crucial role. The MSFF-FE module enhances multi-scale feature fusion while preserving high-frequency components, the FD module focuses on maintaining spatial details during downsampling, and the SAC module ensures semantic alignment across different feature paths. Experimental results on the VisDrone and UAVVaste datasets demonstrate that our method achieves higher accuracy than existing approaches with similar computational costs while maintaining real-time inference speeds. Future work will focus on improving its robustness to noise.

\bibliographystyle{IEEEtran}
\bibliography{ref}
\end{document}